\documentclass{elsarticle}
\usepackage{url}  
\usepackage{times}
\usepackage{xcolor}
\usepackage{soul}
\usepackage[utf8]{inputenc}
\usepackage[small]{caption}
\usepackage{graphicx}  

\usepackage{amsmath}
\usepackage{amssymb}
\usepackage{multirow}

\journal{Journal of Visual Communication and Image Representation}

\begin{document}

\begin{frontmatter}

\title{Exploring Frame Segmentation Networks for Temporal Action Localization}
\author{{Ke Yang, Xiaolong Shen, Peng Qiao, Shijie Li, Dongsheng Li and Yong Dou}}
\address{National Laboratory for Parallel and Distributed Processing\\
College of Computer, National University of Defense Technology\\
Changsha, China\\}

\begin{abstract}
Temporal action localization is an important task of computer vision. Though many methods have been proposed, it still remains an open question how to predict the temporal location of action segments precisely. Most state-of-the-art works train action classifiers on video segments pre-determined by action proposal. However, recent work found that a desirable model should move beyond segment-level and make dense predictions at a fine granularity in time to determine precise temporal boundaries.
In this paper, we propose a Frame Segmentation Network (FSN) that places a temporal CNN on top of the 2D spatial CNNs. Spatial CNNs are responsible for abstracting semantics in spatial dimension while temporal CNN is responsible for introducing temporal context information and performing dense predictions.
The proposed FSN can make dense predictions at frame-level for a video clip using both spatial and temporal context information.
FSN is trained in an end-to-end manner, so the model can be optimized in spatial and temporal domain jointly. We also adapt FSN to use it in weakly supervised scenario (WFSN), where only video level labels are provided when training. Experiment results on public dataset show that FSN achieves superior performance in both frame-level action localization and temporal action localization.
\end{abstract}

\begin{keyword}
Action detection\sep Temporal action localization\sep Convolutional Neural Network
\end{keyword}

\end{frontmatter}

\section{Introduction}
In recent years, temporal action localization has been extensively studied by researchers in computer vision. A lot of works have been tried to solve
this problem \cite{escorcia2016daps,idrees2017thumos,shou2016temporal,yeung2016end,shou2017cdc,lin2017single},
but how to perform temporal action localization precisely is still an open question. Temporal action localization aims to detect action instances in the untrimmed videos, including their temporal boundaries and categories. Most works adopt the detection by classification framework which is widely used in object detection task \cite{girshick2014rich}. First, action segment proposals are generated by action proposal methods or sliding windows. Then various features are extracted on action segment proposals and action classifiers are trained on these extracted features.

A recent work claimed that action prediction at a finer temporal granularity
contributes to more precise temporal action localization results
\cite{shou2017cdc}. This finding encourages us to perform action prediction at
a fine granularity rather than at segment-level. To achieve this goal, there
are some techniques can be adapted: (1) Recurrent Neural Network (RNN); (2) 3D
CNN; (3) 2D CNN; In \cite{yeung2015every}, a Long Short Term Memory (LSTM)
network is proposed to model these temporal relations via multiple input and
output connections. However, it is claimed that RNN will introduce temporal
smoothing that is harmful to precise temporal localization task in
\cite{yeung2016end,shou2017cdc}. In \cite{shou2017cdc}, 3D CNN is reformed to
accomplish this goal. 3D CNN is designed to classify a whole video clip. To
perform frame-level predictions, a Convolutional-De-Convolutional (CDC) layer
is developed to upsample the temporal resolution. However, the model parameters
of 3D CNN increase significantly relative to 2D CNN and 3D CNN is much more
data hungry than 2D CNN \cite{carreira2017quo}. The increase in the number of
parameters makes the computational resource and training time consumption
increase significantly. Meanwhile, there are very few pre-trained 3D CNN models
available \cite{tran2015learning}.

In order to reduce the number of parameters while modeling spatio-temporal information, a valid method is to decompose the spatial and temporal dimensions of 3D CNN. \emph{We propose to combine 2D spatial CNNs and 1D temporal CNN instead of 3D CNN to model spatio-temporal information.} Formally, the 1D temporal CNN is placed on top of the 2D CNNs.
We have also considered separable 3D convolution for space and time, but this operation changes the internal structure of the network and  might invalidate the application of pre-training model weights which is important for relatively small dataset.

In recent years, due to the rapid development of image recognition, 2D CNN has
been developed by leaps and bounds, deeper and deeper networks with stronger
capacity are being proposed \cite{simonyan2015very,he2016deep}. These
state-of-the-art 2D CNN models pre-trained on ImageNet \cite{deng2009imagenet}
can be transferred to action recognition with a small computational and time
cost \cite{wang2016temporal,simonyan2014two,tsn-tpami,wang2015exploring}. 2D
CNN has already been
able to model spatial information successfully. However, 2D CNNs classify each
frame using 2D CNN without consideration of temporal information which is
important for video understanding.
Therefore, we consider stacking a 1D temporal CNN on top of a 2D CNN to model temporal information. As shown in Figure \ref{fig:pipeline}, 2D CNNs take single images as input and model the spatial information. \emph{All these 2D CNNs share the weights.} A 1D temporal CNN is placed on top of the 2D CNN. \emph{The temporal CNN takes a sequence of feature vectors from 2D CNN and outputs dense predictions of each input frame through a single pass.} This forms our Frame Segmentation Network (FSN). FSN allows the model to take multiple video frames as input and output predictions for every input frame. And we can easily control the temporal receptive field size by setting different kernel size and step size for the temporal CNN. FSN can be trained in an end-to-end manner.

Most action localization works belong to strongly supervised method. Weakly supervised temporal action localization has not been studied much. However, the research on weakly supervised action localization is necessary due to a few reasons. First, the training procedure of CNN consumes large amounts of labeled data and complex annotation in this task needs large amount of manual efforts and time. Second, the boundaries of actions are easily confused and annotations are easy to be affected by the subjective factors. Our FSN can be adapted into weakly supervised version through minor changes.

Our contributions can be concluded as follows: (1) Our FSN can make dense predictions with temporal context information and can be trained in an end-to-end manner. (2) FSN achieves competitive results in both per-frame action localization and segment-level action localization. It is also worth noting that FSN can be easily updated by simply changed the 2D CNN to a more powerful one, allowing FSN to benefit from the progress of image classification network.

\section{Related work}

\label{re_work}
\textbf{Action recognition:} For a quite long period of time, the conventional features such as Improved Dense Trajectory Feature (iDTF) \cite{wang2013action} were in a dominant position in the field of action recognition. In recent years, thanks to ImageNet dataset \cite{deng2009imagenet}, Convolutional Neural Networks (CNN) such VGG \cite{simonyan2015very}, ResNet \cite{he2016deep} have gradually been proposed and adapted to perform action recognition, but the performance is still poor since they can only capture appearance information. To model motion information in videos, various two-stream CNNs which take both RGB images and optical flow as input have significantly improve the action recognition performance and surpass the conventional features \cite{wang2016temporal,simonyan2014two}. To explicitly model spatio-temporal feature directly from raw videos, a 3D CNN architecture called C3D is proposed in \cite{tran2015learning}.

\textbf{Temporal action localization:}
A typical framework used in many state-of-the-art temporal action localization systems \cite{oneata2014lear,shou2016temporal} is detection by classification framework, which is borrowed from object detection task. First, various features are extracted on the action segments pre-determined by action proposals. Then action classifiers are trained on these features to classify these action segments. In order to design a model specific to temporal localization, in \cite{richard2016temporal}, statistical length and language modeling are used to represent temporal and contextual structure. In \cite{caba2016fast} a sparse learning framework is proposed to retrieve action segment proposals of high recall.

In recent years, deep neural networks are used widely to improve performance of temporal localization. In \cite{yeung2016end}, a Long Short Term Memory (LSTM)-based agent is trained using REINFORCE to learn  both which frame to look in the next step and when to emit an action segment output.
In \cite{escorcia2016daps}, a LSTM based framework is designed to takes pre-extracted CNN features to output temporal action proposal. In \cite{yeung2015every}, a MultiLSTM network is proposed to model these temporal relations via multiple input and output connections. In \cite{yuan2016temporal}, a Pyramid of Score Distribution Feature (PSDF) is proposed and PSDF is taken as input into the RNN to improve temporal consistency.
A novel Single Shot Action Detector (SSAD) network is proposed to skip the
proposal generation step via directly detecting action instances in untrimmed
video \cite{lin2017single}. In \cite{tran2015learning}, an end-to-end framework
named Segment-CNN (S-CNN) is proposed to perform action localization via
multi-stages 3D CNNs. Convolutional-De-Convolutional (CDC) is proposed to
perform action predictions in every frame in \cite{shou2017cdc} and then the
frame-level action predictions are used to refine the action segment boundaries
from S-CNN to generate more precise segment-level predictions. Visual tracking
is also related to this task \cite{lan2014multi,lan2015joint,lan2018learning},
the tracking results can help the detection.

\textbf{Weakly supervised learning in images and videos:}
Weakly supervised object detection has been studied a lot in the past few years
\cite{bilen2016weakly}. Several works try to adapt these methods to weakly
supervised action understanding
\cite{wang2017untrimmednets,laptev2008learning,bojanowski2014weakly,singh2017hide}.
 These works can be divided into three types according to the used weak
supervision. The first type is movie script, which gives uncertain temporal
annotations of action instances \cite{laptev2008learning}. The second type is
an ordered list of action classes occurring in the videos
\cite{bojanowski2014weakly}. The third type weak supervision only contains
video labels which contain no any order information of the containing action
instances but whether a action category appears in a video
\cite{wang2017untrimmednets,singh2017hide,nguyen2018weakly}. Our weakly
supervised FSN falls
into the third type.

\section{Frame segmentation networks}

\subsection{Motivation of frame segmentation networks}
\begin{figure}[t]
\begin{center}
    \includegraphics[width=1\linewidth]{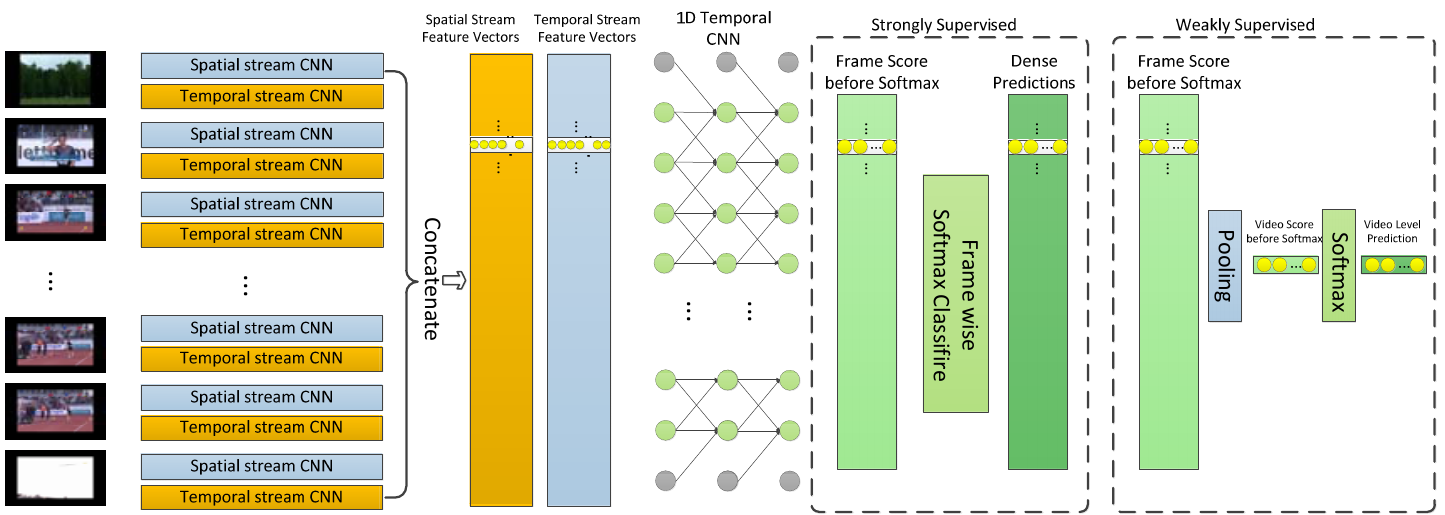}
\end{center}
   \caption{Overall training pipeline of FSN. Our FSN takes a sequence of video frames as input. Then a sequence of feature vectors are extracted for every input video frame. Then the feature vectors sequence are passed to a 1D temporal CNN and output a sequence of scores of the same time length. Next, for strongly supervised FSN, these scores are passed through a frame wise softmax classifier to output the dense action predictions for every input frame; for weakly supervised FSN, the sequence of scores is reduced to a single score vector by pooling operation such as Global Average Pooling (GAP), then the single vector is passed through a softmax classifier to yield a single video level prediction of the whole input video. This figure is for illustrative purpose only, we omit the upsampling layer and the specific structure of temporal CNN. Please zoom in for better viewing.}
   \label{fig:pipeline}
\end{figure}

The 3D CNN is naturally suitable for encoding video data, but the amount of parameters of 3D CNN is too large, making the training process consume a lot of computing resources and time. For example, a 11-layer 3D CNN \cite{tran2015learning} has about 80M parameters while a 50-layer ResNet \cite{he2016deep} only has about 25M parameters. To reduce the model parameter number, we turn to model spatio-temporal information using a combination of 2D spatial CNNs and 1D temporal CNN rather than 3D CNN. At the same time, state-of-the-art 2D CNNs, such as VGG \cite{simonyan2015very} and ResNet \cite{he2016deep}, have been adapted for action recognition task and get the state-of-the-art results \cite{wang2016temporal}, we can use any one of them to initialize our 2D CNN. And with the 1D temporal CNN, we can predict action class scores at the original temporal resolution rather than output a single video-level label.
The training pipeline of FSN is shown in Figure \ref{fig:pipeline}.

\subsection{Feature extraction}
\label{subsection:feat}
First of all, video clip frames are fed into a deep 2D CNN for feature extraction.
A video clip $V$ contains $T$ frames can be divided into $N$ consecutive snippets. We choose the center frame of each snippet $C = {\{c_{i}\}}_{i=1}^N$ to represent the snippet. When snippet's length is 1, the center frame is the snippet itself.

Each center frame is processed by a 2D CNN to extract the representation as $f_{c_i} = \phi(V;c_i) \in \mathbb{R}^D$. Then all $\phi(V;C)$ of the center frames are concatenated together to $f_{con} \in \mathbb{R}^{N \times D}$. Then $f_{con}$ is fed to a 1D temporal CNN.

Our FSN does not depend on the choice of 2D CNN feature extraction model. In our experiments, a two stream network like network for action recognition: Temporal Segment Network (TSN) \cite{wang2016temporal} is investigated due to its superior performance on action recognition task. Feature vectors after global pooling layer are used.

\subsection{Temporal CNN}

Representation of the input video clip $f_{con}$ is a simple concatenation of feature vectors of multiple frames. A desirable model is that can take temporal context information to predict each frame of input video through a single pass.
The temporal CNN enables the entire model to see the temporal context when predict the current frame.
The temporal CNN is a 1D CNN with $L$ layers. Following network architecture design of classic CNN model VGG \cite{simonyan2015very}, all the kernel sizes are set to 3. All kernels' steps are set to 1, since dense predictions needs the model to preserve the time length. Convolutional kernels with step size of 1 raise a problem that the temporal receptive field size will be too small, but too small temporal receptive field is harmful to the prediction precision. To solve this problem, we adapt spatial dilated convolution \cite{yu2015multi} to \emph{temporal dilated convolution} to enlarge the temporal receptive field.  Temporal dilated convolution is used in all but the first layers of temporal CNN. Temporal CNN has three convolution layers except the classification layer. Dilated rate is set to 1,2 and 4 respectively from first to last convolution layers.

\textbf{Bilinear upsampling and classifier.} The output of temporal CNN is $X = {\{x_{i}\}}_{i=1}^N$. $x_{i} \in \mathbb{R}^{K + 1}$ is the score vector of the $i^{th}$ snippet. $N$ is snippet number. $K$ is the action category number. $(K+1)$ dimension score vector corresponds to $K$ action categories and 1 background category. FSN needs to perform frame-level predictions, thus when the snippet length is larger than 1 frame, we need to upsample the number of score vector to the frame number of input video clip. We choose bilinear upsampling which has no parameter following \cite{long2015fully}. $X \in \mathbb{R}^{N \times (K + 1)}$ is upsampled to $X_{up} \in \mathbb{R}^{T \times (K + 1)}$ by $X_{up} = BilinearUpsampling1D(X)$. $T$ is the frame number of input video clip as described in Section \ref{subsection:feat}.
Then the upsampled feature vectors $ X_{up}$ are passed through a frame wise softmax layer as follows:
 \begin{equation}\label{eq:frm_wise_softmax}
  \overline{x}_{up\ i}^j = \frac{exp\left(x_{up\ i}^j\right)}{\sum_{k=1}^{K + 1} exp\left(x_{up\ i}^k\right)}
\end{equation}
where $x_{up\ i}^j$ denotes the $j^{th}$ dimension of the $i^{th}$ input frame's score vector ${x}_{up\ i}$. $\overline{x}_{up\ i}^j$ denotes the $j^{th}$ dimension of the $i^{th}$ input frame's softmax score vector $\overline{x}_{up\ i}$.
We have presented the pipeline of strongly supervised FSN. It is clearly that FSN can be trained in an end-to-end manner.

\subsection{Weakly supervised FSN}
\label{wfsn}
Weakly supervised temporal action localization methods can only see video-level label when training, and they need to output instance-level labels include temporal boundaries and categories of each detected action instance.
Weakly supervised FSN takes $X = {\{x_{i}\}}_{i=1}^N$ as input. First, $X$ is passed through temporal pooling layer as follows:

 \begin{equation}\label{eq:pooling}
 x_{pool} = TemporalPooling({\{x_{i}\}}_{i=1}^N)
\end{equation}

where $x_{pool} \in \mathbb{R}^{K}$ represents the score vector for the whole video. For weakly supervised version, $x_{pool}$ and $x_{i}$ all are $K$ dimension feature vectors. $K$ is the number of action categories. Background category
is not included in the weakly supervised scenario. Then the score vector $x_{pool}$ of the whole video is passed through a standard softmax layer to as follows:
 \begin{equation}\label{eq:softmax}
  \overline{x}^j = \frac{exp\left(x_{pool}^j\right)}{\sum_{k=1}^{K} exp\left(x_{pool}^k\right)}\end{equation}
where $\overline{x}^j$ denotes the $j^{th}$ dimension of the video's softmax score vector.

\textbf{Temporal pooling.} Given the prior works \cite{zhou2016learning} in weakly supervised object detection, Global Average Pooling (GAP) or Global Max Pooling (GMP) can be used for the temporal pooling in Equation \ref{eq:pooling}. In \cite{zhou2016learning}, the intuitive difference between GAP
and GMP is discussed.
GAP encourages the network to identify the full body of the instance while GMP is considered to be able to force the network to learn the most discriminating position. It is found in \cite{zhou2016learning} that GAP performs better than GMP for weakly supervised localization. However, GAP might not be suitable for weakly supervised temporal action localization, since action instances only occupy a very small part of the whole video, thus GAP is very likely to force CNN to learn both action instances and parts of correlated or closely situated background. We verify this experimentally on THUMOS'14 dataset in Section \ref{disc}: GMP outperforms GAP for weakly supervised temporal action localization. We use GMP in all weakly supervised experiments except in Section \ref{disc}.

\subsection{Model training and prediction}
\label{subsec:model_train_prediction}
\textbf{Training data construction of FSN.} Training data of strongly supervised FSN consists of video clips with temporal length $T$ frames. $T$ can be an arbitrary value since temporal CNN is a 1D fully convolutional network. Following \cite{shou2017cdc}, and considering that the temporal stream network of Temporal Segment Network \cite{wang2016temporal} takes 5 adjacent optical flows as input, we set $T$ to 35 frames which is a multiple of 5. Therefore, the snippet number $N$ for FSN is $35/5=7$. We slide temporal window of length $T$ on the videos and only keep the segments include at least 5 frames belongs to action instances. We also re-sample the segments to get a balance training dataset.

\textbf{Training data construction of WFSN.} In weakly supervised scenario, we can only obtain video level labels which indicate which action happens in the videos. So WFSN needs to take the whole video as input. We divide a whole video to $M$ segments with equal length. Then we randomly choose a frame in the each segment to represent that segment. Then we use 2D CNN extract feature on each segments to form the feature vector $X = {\{x_{i}\}}_{i=1}^M$. $M$ is set to 100.

\textbf{Model training}. We implement FSN based on Keras \cite{chollet2015keras} and Temporal Segment Network (TSN) \cite{wang2016temporal}. For experiments on THUMOS'14, we first finetune TSN on UCF101. For FSN, the \emph{frame wise} cross-entropy loss ${\mathcal{L}}$ is as follows:
\begin{equation}\label{eq:Loss}
  {\mathcal{L}} = \frac{1}{B} \sum_{b=1}^{B} \sum_{t=1}^{L} \sum_{k=1}^{K + 1} \left(-y^{(k)}_{n}[t]\log \left(\frac{\exp\left(O^{(k)}_{n}[t]\right)}{\sum_{j=1}^{K + 1} \exp\left(O^{(j)}_{n}[t]\right)}\right)\right)
\end{equation}
where $B$ denotes batch size, $L$ denotes video length, $K$ denotes the action category. We use Stochastic Gradient Descent (SGD) to train FSN network. We train all layers of FSN with learning rate 0.0001 and mini-batch size 12. We set momentum to 0.9 and weight decay to 0.0005. Training iteration is about 60000.

For WFSN, we use the standard cross-entropy loss. Optimization method is similar to FSN. We freeze the first few convolutional layers of WFSN since the training data size for WFSN is too small, which can easily lead to overfitting.

\textbf{FSN model prediction}.
First, we introduce frame-level action predictions. During test, we slide the FSN on the whole untrimmed videos without temporal overlapping. We get action predictions of every frame in the test set.

Then, we introduce segment-level action localization predictions. After frame-level action predictions, we have the predictions of all frames. Then we can generate segment-level action predictions by grouping frame-level action scores. First, we take threshold processing on classification scores and we get a string of "0" and "1" (0 stands for background frame, and 1 inversely). Then we group adjacent frames of "1" to get segment-level results. Thresholds are uniformly selected from 0 to 1 with an interval of 0.1.

\textbf{WFSN model prediction}.
For weakly supervised FSN, we remove the pooling layer and slide the last layer on the feature vector to obtain dense predictions. Other details are the same as strongly supervised FSN.

\section{Experiments}
\subsection{Dataset}
We evaluate FSN network and weakly supervised FSN (WFSN) on the challenging dataset THUMOS'14 \cite{idrees2017thumos}.

\textbf{THUMOS'14 dataset.} THUMOS'14 has 101 action classes. Training set is directly taken from UCF101 dataset \cite{soomro2012ucf101}. Validation set consists of 1010 untrimmed videos. Test set consists of 1574 untrimmed videos. Temporal action detection task in THUMOS'14 challenge is dedicated to localize the action instances in untrimmed video and involves 20 action classes. And there are 200 videos in validation set and 213 videos in test set that contain the action instances of these 20 classes. we train FSN on the 200 validation videos and test on the 213 test videos. On THUMOS'14, We evaluate strongly supervised FSN network on both frame-level action localization and segment-level action localization tasks and evaluate weakly supervised FSN network on segment-level action localization task. For weakly supervised action localization, state-of-the-art work UntrimmedNets \cite{wang2017untrimmednets} also use UCF101 as training data. To have a fair comparison, we first finetune the TSN on UCF101 dataset for weakly supervised experiments.
\begin{table}[t]

\footnotesize
\begin{center}
\begin{tabular}{c|c}

\multicolumn{1}{c|}{Method} & \multicolumn{1}{c}{mAP}\\
\hline
\multicolumn{1}{c|}{Single-frame CNN\cite{simonyan2015very}} & \multicolumn{1}{c}{34.7} \\

\multicolumn{1}{c|}{Two-stream CNN\cite{simonyan2014two}} & \multicolumn{1}{c}{36.2} \\

\multicolumn{1}{c|}{LSTM\cite{donahue2015long}} & \multicolumn{1}{c}{39.3} \\

\multicolumn{1}{c|}{MultiLSTM\cite{yeung2015every}} & \multicolumn{1}{c}{41.3} \\

\multicolumn{1}{c|}{CDC\cite{shou2017cdc}} & \multicolumn{1}{c}{44.4}\\

\hline
\multicolumn{1}{c|}{\textbf{FSN} RGB} & \multicolumn{1}{c}{47.5}\\

\multicolumn{1}{c|}{\textbf{FSN} Flow} & \multicolumn{1}{c}{41.4}\\
\multicolumn{1}{c|}{\textbf{FSN}} & \multicolumn{1}{c}{\textbf{53.5}}\\

\end{tabular}
\end{center}
\caption{Frame-level action localization mAP on THUMOS'14.}
\label{table:frm_level}
\end{table}

\subsection{Frame-level action localization}\label{Eval_frm}

First, we evaluate FSN network in predicting action labels for every frame in the whole video. This task can take multiple frames as input to take into account temporal information. Following \cite{yeung2015every}, we evaluate frame-level prediction as a retrieval problem. For each action class, we rank all the frames in the test set by their confidence scores and compute Average Precision (AP) for this class. And mean AP (mAP) is computed by average the AP of 20 action classes.

In Table \ref{table:frm_level}, we compare our FSN network with state-of-the-art methods. All the results are quoted from \cite{yeung2015every,shou2017cdc}. Single-frame CNN stands for frame-level VGG-16 2D CNN model proposed by \cite{simonyan2015very}.
Two-stream CNN \cite{simonyan2014two} is the frame-level 2D CNN model which
takes both optical flow and RGB images as input. LSTM stands for the 2D CNN + LSTM model \cite{donahue2015long}. MultiLSTM represents the LSTM with
temporal attention mechanism \cite{yeung2015every}. CDC denotes the
convolutional-de-convolutional network proposed in \cite{shou2017cdc}.
We denote our FSN network as \textbf{FSN}. Single-frame CNN only takes into
account appearance information in a single frame. Two-stream CNN takes both
appearance information in a single frame and motion information from six
adjacent frames as input. LSTM and MultiLSTM can utilize temporal information
to make frame-level predictions. CDC is based on 3D CNN, can model spatio-temporal
information and make dense predictions by upsampling. Our FSN use 2D CNN to abstract spatial semantics, and use a temporal CNN to pursue temporal context information for dense predictions for each input frame. FSN achieves significant performance improvement relative to other methods.
 We also report performance of each single stream network in Table \ref{table:frm_level}.

\begin{table*}[t]
\footnotesize
\begin{center}
\begin{tabular}{c|p{30cm}cccccccccccccccccccccccccccccccc}
\multicolumn{1}{c|}{IoU threshold} & \multicolumn{1}{c}{0.3} & \multicolumn{1}{c}{0.4} & \multicolumn{1}{c}{0.5} & \multicolumn{1}{c}{0.6} & \multicolumn{1}{c}{0.7} \\
\hline
\multicolumn{1}{c|}{Wang et al.\cite{wang2014action}} & \multicolumn{1}{c}{14.6} & \multicolumn{1}{c}{12.1} & \multicolumn{1}{c}{8.5} & \multicolumn{1}{c}{4.7} & \multicolumn{1}{c}{1.5} \\
\multicolumn{1}{c|}{Heilbron et al.\cite{caba2016fast}} & \multicolumn{1}{c}{-} & \multicolumn{1}{c}{-} & \multicolumn{1}{c}{13.5} & \multicolumn{1}{c}{-} & \multicolumn{1}{c}{-} \\
\multicolumn{1}{c|}{Escorcia et al.\cite{escorcia2016daps}} & \multicolumn{1}{c}{-} & \multicolumn{1}{c}{-} & \multicolumn{1}{c}{13.9} & \multicolumn{1}{c}{} & \multicolumn{1}{c}{-} \\
\multicolumn{1}{c|}{Oneata et al.\cite{oneata2014lear}} & \multicolumn{1}{c}{28.8} & \multicolumn{1}{c}{21.8} & \multicolumn{1}{c}{15.0} & \multicolumn{1}{c}{8.5} & \multicolumn{1}{c}{3.2} \\
\multicolumn{1}{c|}{Richard and Gall\cite{richard2016temporal}} & \multicolumn{1}{c}{30.0} & \multicolumn{1}{c}{23.2} & \multicolumn{1}{c}{15.2} & \multicolumn{1}{c}{-} & \multicolumn{1}{c}{-} \\
\multicolumn{1}{c|}{Yeung et al.\cite{yeung2016end}} & \multicolumn{1}{c}{36.0} & \multicolumn{1}{c}{26.4} & \multicolumn{1}{c}{17.1} & \multicolumn{1}{c}{-} & \multicolumn{1}{c}{-} \\
\multicolumn{1}{c|}{Yuan et al.\cite{yuan2016temporal}} & \multicolumn{1}{c}{33.6} & \multicolumn{1}{c}{26.1} & \multicolumn{1}{c}{18.8} & \multicolumn{1}{c}{-} & \multicolumn{1}{c}{-} \\
\multicolumn{1}{c|}{S-CNN\cite{shou2016temporal}} & \multicolumn{1}{c}{36.3} & \multicolumn{1}{c}{28.7} & \multicolumn{1}{c}{19.0} & \multicolumn{1}{c}{10.3} & \multicolumn{1}{c}{5.3} \\
\multicolumn{1}{c|}{CDC\cite{shou2017cdc} + S-CNN\cite{shou2016temporal}} & \multicolumn{1}{c}{40.1} & \multicolumn{1}{c}{29.4} & \multicolumn{1}{c}{23.3} & \multicolumn{1}{c}{13.1} & \multicolumn{1}{c}{7.9} \\
\multicolumn{1}{c|}{SSAD\cite{lin2017single}} & \multicolumn{1}{c}{43.0} & \multicolumn{1}{c}{35.0} & \multicolumn{1}{c}{24.6} & \multicolumn{1}{c}{15.4} & \multicolumn{1}{c}{7.7} \\
\multicolumn{1}{c|}{TPC\cite{yang2018exploring}} & \multicolumn{1}{c}{44.1} & \multicolumn{1}{c}{ 37.1} & \multicolumn{1}{c}{28.2} & \multicolumn{1}{c}{20.6} & \multicolumn{1}{c}{12.7} \\
\hline
\multicolumn{1}{c|}{FSN RGB} & \multicolumn{1}{c}{40.7} &
\multicolumn{1}{c}{33.6} & \multicolumn{1}{c}{24.9} &
\multicolumn{1}{c}{17.5} & \multicolumn{1}{c}{10.6} \\
\multicolumn{1}{c|}{FSN Flow} & \multicolumn{1}{c}{36.4} & \multicolumn{1}{c}{28.0} &
\multicolumn{1}{c}{20.0} & \multicolumn{1}{c}{11.9} & \multicolumn{1}{c}{6.2} \\
\multicolumn{1}{c|}{\textbf{FSN}} & \multicolumn{1}{c}{\textbf{51.8}} & \multicolumn{1}{c}{\textbf{41.5}} & \multicolumn{1}{c}{\textbf{32.1}} & \multicolumn{1}{c}{\textbf{22.9}} & \multicolumn{1}{c}{\textbf{14.7}} \\
\end{tabular}
\end{center}
\caption{Segment-level action localization mAP on THUMOS'14. IoU threshold values
are ranged from 0.3 to 0.7. '-' in the table indicates that results of that IoU
value are not available in the corresponding papers. Some of the results were not reported
in the published papers, we contacted the authors for these results.}
\label{table:FSN}
\end{table*}

Given frame-level action predictions, we can get segment-level action localization results using various strategies. As described in Section \ref{subsec:model_train_prediction}, we use multiple threshold frame grouping method to obtain the segment-level localization results. Finally, we perform post-processing steps such as non-maximus suppression (NMS). NMS IoU threshold is 0.1 lower than the IoU threshold used during the evaluation. We evaluate our model on THUMOS'14 dataset.

We perform evaluation using mAP as frame-level action localization evaluation. For each action class, we rank all the predicted segments by their confidence results and calculate the AP using official evaluation code. One prediction is correct when its temporal overlap intersection-over-union (IoU) with a ground truth action segment is higher than the threshold, so evaluation under various IoU threshold is necessary. We evaluate our model under IoU threshold from 0.3 to 0.7 following most works \cite{shou2016temporal,shou2017cdc,yeung2016end}.
Results are shown in Table \ref{table:FSN}, our model denoted as \textbf{FSN} achieves competitive results compared to current state-of-the-art results.

\subsection{Weakly supervised action localization}
\label{wsal}
For weakly supervised FSN, the temporal pooling layer is removed during test. Thus, we can obtain a similar prediction process as strongly supervised FSN. We group the adjacent frames by taking threshold on softmax score and grouping the frames whose scores are larger than threshold as described in Section \ref{subsec:model_train_prediction}.

\textbf{Experiment results on THUMOS'14}. We show the weakly supervised segment-level action localization results on THUMOS'14 in Table \ref{table:weakFSN_th}. Since previous methods using weakly supervision are evaluated under the IoU from 0.1 to 0.5 \cite{singh2017hide,wang2017untrimmednets},
we also report performance under the same IoU thresholds. In Table \ref{table:weakFSN_th}, UntrimmedNet \cite{wang2017untrimmednets} is an end-to-end action localization network. Attention vector is used to perform weakly supervised temporal action localization. Video-Has \cite{singh2017hide} uses an random input mask trick to train a weakly supervised action localization network. We noticed that results of Video-Has \cite{singh2017hide} in Table \ref{table:weakFSN_th} is on validation set of THUMOS'14 dataset. They assume that they know the ground truth video level labels. They only need to locate the temporal position of the action instances. Results in Table \ref{table:weakFSN_th} suggest that WFSN's performance is superior to other weakly supervised methods and is comparable
to several strongly supervised methods in Table \ref{table:FSN}.
\subsection{Discussions}
\label{disc}
In this subsection, for purpose of saving time, we only use RGB stream network of TSN.

\begin{table}[t]
\footnotesize
\begin{center}
\begin{tabular}{c|ccccccccccc}
\multicolumn{1}{c|}{IoU threshold} & \multicolumn{1}{c}{0.1} & \multicolumn{1}{c}{0.2} & \multicolumn{1}{c}{0.3} & \multicolumn{1}{c}{0.4} & \multicolumn{1}{c}{0.5} \\
\hline

{UntrimmedNet} & \multirow{2}*{44.4} & \multirow{2}*{37.7} & \multirow{2}*{28.2} & \multirow{2}*{21.1} & \multirow{2}*{13.7} \\
\cite{wang2017untrimmednets} & {} & {} & {} & {} & {} \\
\hline
{Video-Has} & \multirow{2}*{36.4} & \multirow{2}*{27.8} & \multirow{2}*{19.5} & \multirow{2}*{12.7} & \multirow{2}*{6.8} \\
\cite{singh2017hide} & {} & {} & {} & {} & {} \\

\hline
\multicolumn{1}{c|}{\textbf{WFSN}} & \multicolumn{1}{c}{\textbf{54.9}} & \multicolumn{1}{c}{\textbf{48.1}} & \multicolumn{1}{c}{\textbf{38.9}} & \multicolumn{1}{c}{\textbf{27.8}} & \multicolumn{1}{c}{\textbf{16.8}} \\
\end{tabular}
\end{center}
\caption{Weakly supervised segment-level action localization mAP on THUMOS'14. IoU threshold values are ranged from 0.1 to 0.5.}
\label{table:weakFSN_th}
\end{table}

\textbf{The necessity of modeling temporal information.} \emph{TSN can also perform dense predictions via proper adaptation, do we really need the temporal CNN to model temporal information?} To answer this question, we design an ablation experiment. The temporal CNN is replaced by a 1D convolution layer of which both the kernel size and
step size are 1. The number of the convolutional layer's output nodes
is $K + 1$, $K$ is the number of action categories. The convolutional layer is followed by a
 upsampling layer and frame wise softmax layer that are the same as FSN. We denote
 this model as \textbf{FSN w/o Temporal CNN}. FSN w/o Temporal CNN can only see a single frame when
 it predict a frame, since the kernel size is 1. We train  FSN w/o Temporal CNN and our FSN with
 the same training set and evaluate them using the same metric. RGB FSN w/o Temporal CNN's frame-level performance is 42.5 (mAP), while our RGB FSN's is 47.5 as shown in Table \ref{table:frm_level}. The segment-level experiment
 results are shown in Table \ref{table:ness_of_temporalCNN}. These results are reasonable since in temporal action localization task, only looking at current frame is very likely to make false predictions. A bigger temporal
 receptive field size helps improve performance. For example, for the $CleanAndJerk$
 action category, you might make false positive predictions when the jerk failed.

\begin{table}[t]

\footnotesize
\begin{center}
\begin{tabular}{c|ccccccccccc}
\multicolumn{1}{c|}{IoU threshold} & \multicolumn{1}{c}{0.3} & \multicolumn{1}{c}{0.4} & \multicolumn{1}{c}{0.5} & \multicolumn{1}{c}{0.6} & \multicolumn{1}{c}{0.7} \\
\hline
{RGB FSN} & \multirow{2}*{38.2} & \multirow{2}*{27.6} & \multirow{2}*{19.7} & \multirow{2}*{12.2} & \multirow{2}*{6.9} \\
{w/o temporal CNN} & {} & {} & {} & {} & {} \\
\hline
\multicolumn{1}{c|}{\textbf{RGB FSN}} & \multicolumn{1}{c}{\textbf{40.7}} & \multicolumn{1}{c}{\textbf{33.6}} & \multicolumn{1}{c}{\textbf{24.9}} & \multicolumn{1}{c}{\textbf{17.5}} & \multicolumn{1}{c}{\textbf{10.6}} \\
\hline
\end{tabular}
\end{center}
\caption{Strongly supervised segment-level action localization mAP on THUMOS'14. IoU threshold values are ranged from 0.3 to 0.7.}
\label{table:ness_of_temporalCNN}
\end{table}
\textbf{Temporal pooling method.} In the Section \ref{wsal}, we reported the performance of WFSN with GMP on THUMOS'14. Now, we carry out experiments on the temporal pooling method. Results are shown in Table \ref{table:pooling}. We only report the performance of RGB stream network. Experiments show that GMP indeed performs better than GAP for weakly supervised temporal action localization, confirming our  intuitive analysis in Section \ref{wfsn}.
\begin{table}[t]

\footnotesize
\begin{center}
\begin{tabular}{c|ccccc|c|ccccc}
\multicolumn{1}{c|}{IoU threshold} & \multicolumn{1}{c}{0.1} & \multicolumn{1}{c}{0.2} & \multicolumn{1}{c}{0.3} & \multicolumn{1}{c}{0.4} & \multicolumn{1}{c}{0.5} \\
\hline
\multicolumn{1}{c|}{RGB WFSN w GAP} & \multicolumn{1}{c}{38.0} & \multicolumn{1}{c}{32.5} & \multicolumn{1}{c}{25.0} & \multicolumn{1}{c}{17.0} & \multicolumn{1}{c}{9.8} \\
\hline
\multicolumn{1}{c|}{\textbf{RGB WFSN w GMP}} & \multicolumn{1}{c}{\textbf{44.1}} & \multicolumn{1}{c}{\textbf{39.2}} & \multicolumn{1}{c}{\textbf{29.6}} & \multicolumn{1}{c}{\textbf{20.7}} & \multicolumn{1}{c}{\textbf{11.6}} \\
\end{tabular}
\end{center}
\caption{Weakly supervised segment-level action localization mAP on THUMOS'14. IoU threshold values are ranged from 0.1 to 0.5.}
\label{table:pooling}
\end{table}

\subsection{Qualitative results}
\begin{figure}[htbp]
\begin{center}
   \includegraphics[width=1\linewidth]{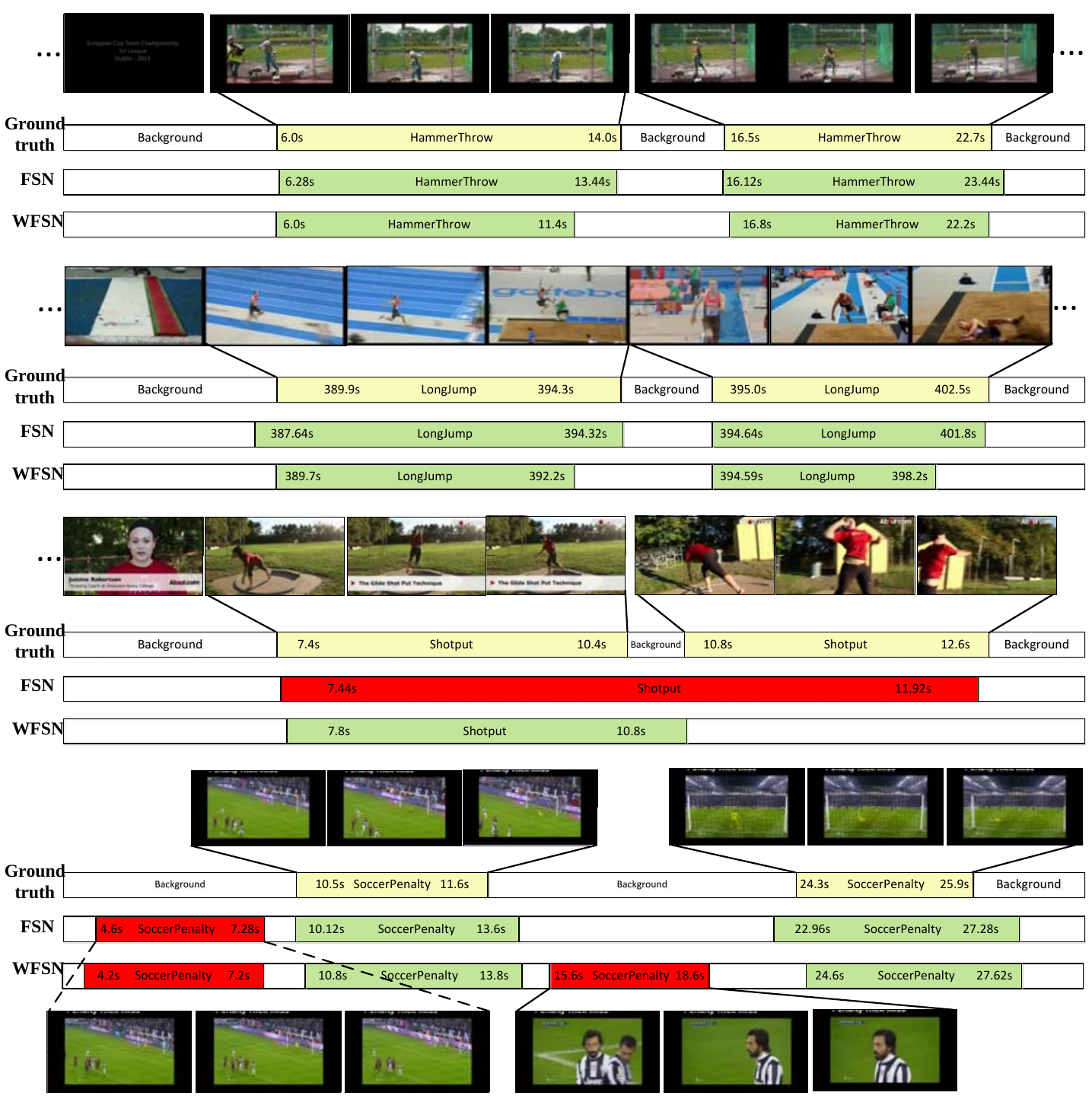}
\end{center}
   \caption{Visualization of temporal localization results FSN and WFSN. The above first two examples show a few successful cases. The last two examples show a few failed cases. In order to have a good display of results, we did not draw the timeline strictly according to the ratio. Please zoom in for better viewing.}
   \label{fig:visualize}
\end{figure}

We show qualitative results of both of FSN and WFSN on segment-level localization
results in Figure \ref{fig:visualize}. The first two examples are two successful cases.
The last two examples includes a few failure case. In the third example, the prediction  of FSN merges
two instances. In this case, these two instances are too close in time and the time interval is too small.
In the fourth example, the predictions of both FSN and WFSN have false positive instances. The common false positive
instance is that soccer penalty shooter is walking to the penalty position. This might be because the receptive field
size is still not big enough. This encourage us to exploit other techniques such as multiple resolution in future. The second
false positive instance of WFSN is because there are temporal boundaries are not given when training. The model might
predict a video clip as $SoccerPenalty$ if there are lawn and soccer players, since $SoccerPenalty$ action is always accompanied
by the scenes of lawn and soccer players. This indicates that we still need to use a small amount of annotated data.
Another interesting phenomenon is that the predicted segments of the WFSN are generally shorter in time than those of the FSN. This phenomenon also exists in weakly supervised object detection task \cite{bilen2016weakly}. This can be explained by the fact that some parts of an action are much more discriminative than other parts, such as the high jumper crossing the bar is more discriminative than the high jumper's run-up. This also reminds us that
we might need to incorporate additional cue in the model to try to learn the concept of "whole action"  as \cite{bilen2016weakly} claimed.
\section{Conclusion}
In this paper, we proposed a Frame Segmentation Network (FSN) that model spatio-temporal information using a combination of 2D spatial CNNs and a 1D temporal CNN for precise temporal action localization. Our FSN achieves competitive performance on both frame-level and segment-level action localization. We also adapt our FSN to a weakly supervised version which only needs video-level annotations. Performance of weakly supervised FSN is even comparable to many strongly supervised methods.
\section{Acknowledgements}
This work was supported by the National Basic Research Program of
China (973) under Grant No.2014CB340303 and the National Natural Science
Foundation of China under Grants U1435219, 61402507 and 61572515.

\section*{References}

\bibliography{YJVCI_2464}

\end{document}